\documentclass[1p,12pt]{elsarticle}

\usepackage{amssymb}

\usepackage{amsthm}

\usepackage{lineno}
\usepackage{gensymb}
\usepackage{graphicx}
\usepackage{adjustbox}
\usepackage{cleveref}
\usepackage [english]{babel}
\usepackage [autostyle, english = american]{csquotes}
\usepackage{caption}
\usepackage{subcaption}
\usepackage{todonotes}
\usepackage{tikz}
\usepackage{soul}
\usepackage{array}
\usepackage{booktabs}
\usepackage{rotating}

\MakeOuterQuote{"}

\journal{Energy}

\begin{document}

\begin{frontmatter}

\title{Effects of spatiotemporal correlations in wind data on neural network-based wind predictions}

\author{Heesoo Shin \fnref{label1}}
\ead{hs_Shin@inha.edu}
\author{Mario Rüttgers \fnref{label2,label3,label4}}
\ead{m.ruettgers@aia.rwth-aachen.de}
\author{Sangseung Lee \corref{cor1}\fnref{label1}}
\ead{sanseunglee@inha.ac.kr}

\cortext[cor1]{Coressponding author}

\affiliation[label1]{organization={Department of Mechanical Engineering},
            address line={Inha University},
            city ={Incheon},
            postcode ={22212},
            country={Republic of Korea}}
\affiliation[label2]{organization={Institute of Aerodynamics and Chair of Fluid Mechanics (AIA)},
            address line={RWTH Aachen University},
            city ={Aachen},
            postcode ={52062},
            country={Germany}}
\affiliation[label3]{organization={Jülich Supercomputing Centre (JSC)},
            addressline={Forschungszentrum Jülich GmbH},
            city ={Jülich},
            postcode ={52425},
            country={Germany}}
\affiliation[label4]{organization={Jülich Aachen Research Alliance - Center for Simulation and Data Science (JARA-CSD)},
            city ={Aachen},
            postcode ={52074},
            country={Germany}}

\begin{abstract}
This paper investigates the influence of incorporating spatiotemporal wind data on the performance of wind forecasting neural networks. While previous studies have shown that including spatial data enhances the accuracy of such models, limited research has explored the impact of different spatial and temporal scales of input wind data on the learnability of neural network models. In this study, convolutional neural networks (CNNs) are employed and trained using various scales of spatiotemporal wind data. The research demonstrates that using spatiotemporally correlated data from the surrounding area and past time steps for training a CNN favorably affects the predictive performance of the model. The study proposes correlation analyses, including autocorrelation and Pearson correlation analyses, to unveil the influence of spatiotemporal wind characteristics on the predictive performance of different CNN models. The spatiotemporal correlations and performances of CNN models are investigated in three regions: Korea, the USA, and the UK. The findings reveal that regions with smaller deviations of autocorrelation coefficients (ACC) are more favorable for CNNs to learn the regional and seasonal wind characteristics. Specifically, the regions of Korea, the USA, and the UK exhibit maximum standard deviations of ACCs of 0.100, 0.043, and 0.023, respectively. The CNNs wind prediction performances follow the reverse order of the regions: UK, USA, and Korea. This highlights the significant impact of regional and seasonal wind conditions on the performance of the prediction models.

\end{abstract}

\begin{graphicalabstract}
\includegraphics[width=\linewidth]{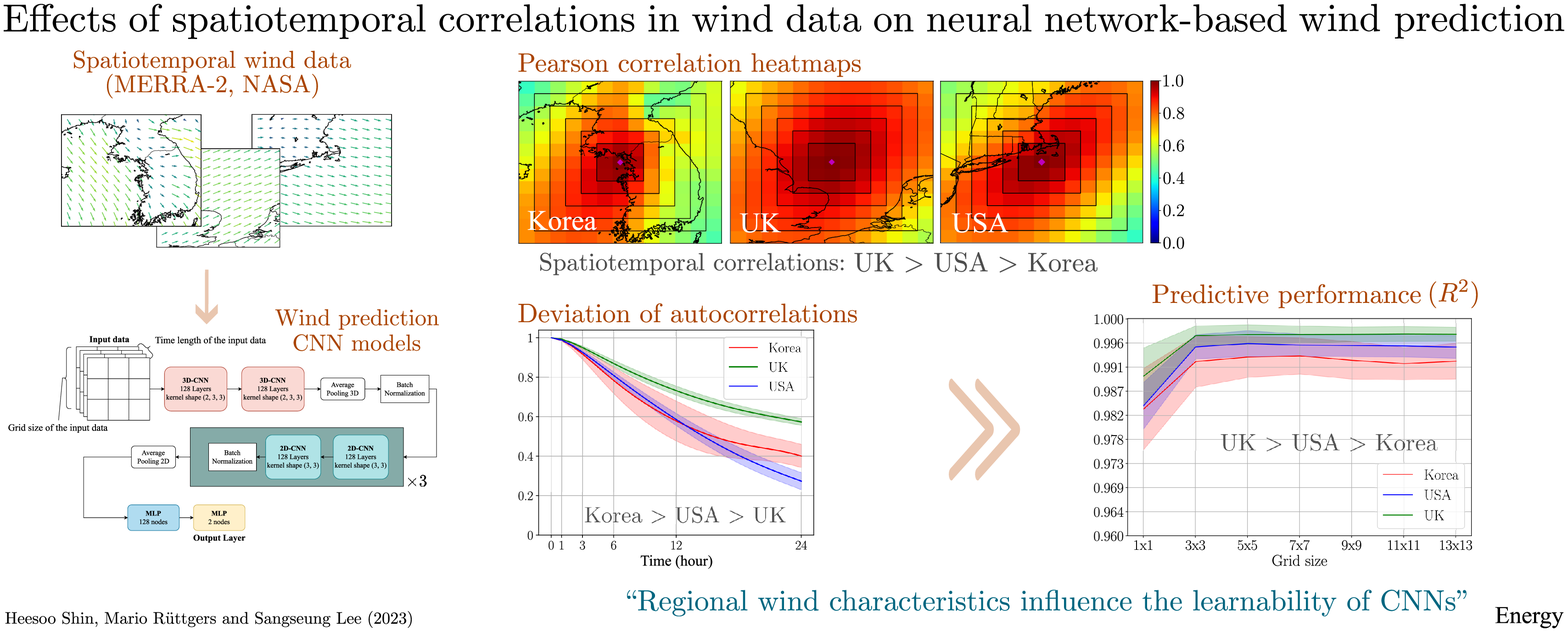}
\end{graphicalabstract}

\begin{highlights}
\item 3D CNN-based wind velocity prediction.

\item Estimation of CNN learnability through spatiotemporal wind correlation analysis.

\item Discovery of influence of local geometric and seasonal wind on CNN prediction.

\item The proposed correlation analysis can aid in selecting yaw-control wind farm sites.

\end{highlights}


\end{frontmatter}

\begin{linenumbers}

\section{Introduction}
\label{intro}
With rising concerns regarding global warming and energy security, there is an increasing demand for renewable energy sources, such as wind energy~\cite{gwec2022}. Wind turbines convert the kinetic energy of atmospheric flow into electrical energy. The maximum power generation of wind turbines depends significantly on the alignment of the turbine nacelle with the surrounding flow~\cite{yang2021review,song2018maximum}. Yaw control systems have been proposed to align wind turbines with the wind direction~\cite{kim2014yaw, wang2021review, hure2015optimal}. One of the most common methods is to use sensors installed at the rear of the turbine to align the turbine with the wind direction. However, the wake effect caused by the rotating blades can lead to deviations in the wind velocity measured by sensors from the actual wind speed~\cite{wang2021review}. 
Therefore, accurately predicting the wind direction remains a significant challenge.
Researchers are investigating methods for accurately predicting the wind direction to enable effective yaw control. 

Recently, neural networks have shown promising results in addressing atmospheric flow problems, such as typhoon prediction~\cite{ruttgers2019prediction, ruttgers2022prediction}. Neural-network-based wind predictions have also been investigated by various researchers~\cite{harbola2019one, ramasamy2015wind, jiajun2020ultra}. These studies primarily involved the training of neural networks using wind data from a single point.
Although wind turbines are installed at specific locations, the wind itself is not a localized phenomenon. This is influenced by macroscopic systems and global parameters. In a study by~\citet{hong2020day}, spatiotemporal wind data were utilized in a 2D-Convolutional neural network (CNN) model to predict wind at a specific location. The training data were sourced from multiple locations, including nearby wind farms in close proximity. The 2D-CNN model outperformed long short-term memory (LSTM) and 1D-CNN models that used data from only one wind farm, demonstrating the importance of utilizing spatiotemporal data for wind prediction.

In a study by \citet{higashiyama2018feature}, the impact of surrounding spatial data on wind power generation was investigated. The dataset used in the study consisted of numerical weather data collected from $50\times 50 = 2,500$ points surrounding a single targeted wind power plant in the Tohoku region with a time resolution of $30$ minutes and regular horizontal spacing of $5$ km. A 3D-CNN model was employed to analyze the spatiotemporal data, which is capable of learning spatial and temporal features concurrently. The results of this study showed that the 3D-CNN model outperformed the 2D-CNN model. In addition, \citet{zhu2021wind} utilized a Fully 3D-CNN model to predict wind speed using wind data collected from $36$ individual wind turbines on a wind farm located in China with a time interval of one day. Because the turbines are located in close proximity to one another, the data collected by each turbine can be considered spatially correlated. Their Fully 3D-CNN model was reported to have superior performance compared to two statistical models (the persistence (PR) method and vector autoregression (VAR)) and three neural networks (LSTM, CNN-LSTM, and CNN-gate recurrent unit (GRU)).

Previous studies have demonstrated that the incorporation of spatial data can improve the accuracy of wind prediction models. However, not enough research has been conducted on the physical mechanisms underlying this improvement, as well as the effect of time intervals on input data in such models. Therefore, this study aims to address these gaps by investigating the impact of spatiotemporal wind data on CNN-based wind predictions. The objective is to elucidate the influence of regional and seasonal wind flow factors on the learning capabilities of CNNs. In particular, the role of spatiotemporal wind data in enhancing the performance of CNN models for wind prediction is analyzed. This study examines the effect of varying the input spatial area and time intervals on the model’s predictive capabilities and investigates the influence of regional and seasonal wind flow patterns on predictive performances in different types of CNN models.

The remainder of this paper is organized as follows. In Section~\ref{data_des}, a detailed description of the utilized wind data is provided, including the processing steps required to transform the data into a suitable form for a CNN. Section~\ref{method} outlines the methodological approach used to adjust the wind data to a wind turbine's height (Section~\ref{sec:abl}) and describes the proposed CNN's architecture, which is a combination of 2D and 3D CNNs, and training process (Section~\ref{sec:ann}). In Section~\ref{results}, the performance of the proposed CNN model under varying spatiotemporal input data is analyzed, and the seasonal and regional factors that influence the network's predictive capability are highlighted. In addition, the proposed model was compared with CNN models from other studies. Finally, Section~\ref{conclusions} summarizes the key findings.

\section{Data description}
\label{data_des}
\begin{figure}
    \centering
            \includegraphics[width = 1 \linewidth]{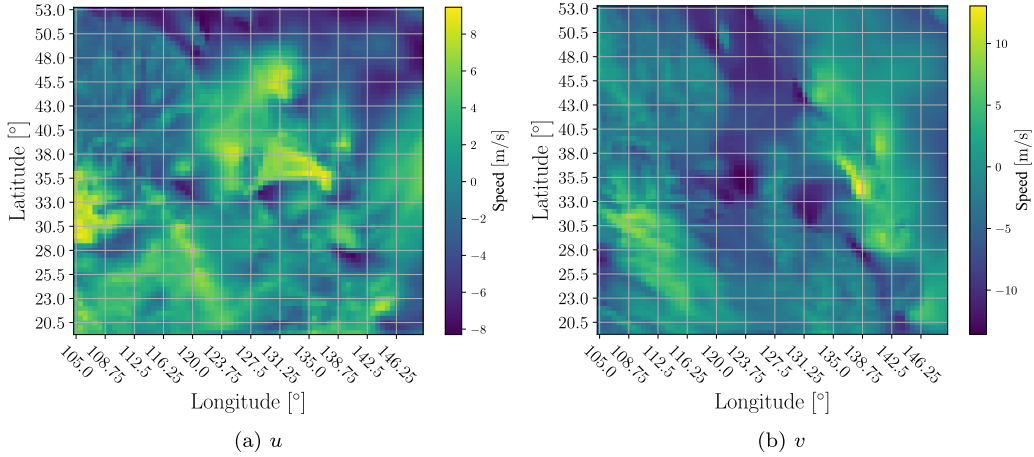}
    \caption{$u$ and $v$ components of wind velocity at an altitude of 50 m in South Korea on January 1, 2012 at 00:00.}
    \label{fig:uvmap}
\end{figure}

This study utilizes the Modern-Era Retrospective Analysis for Research and Applications version 2 (MERRA-2) dataset provided by the National Aeronautics and Space Administration (NASA)\cite{merra-2}. The dataset has a time interval of one hour with hourly averaged values. It was rearranged in a grid format in which each grid point was assigned to the corresponding latitude and longitude coordinates. The grid was constructed at constant intervals in each latitudinal and longitudinal direction, forming a rectangular grid structure (refer to Figure~\ref{fig:uvmap}).

\begin{figure}
  \centering
    \includegraphics[width=\linewidth]{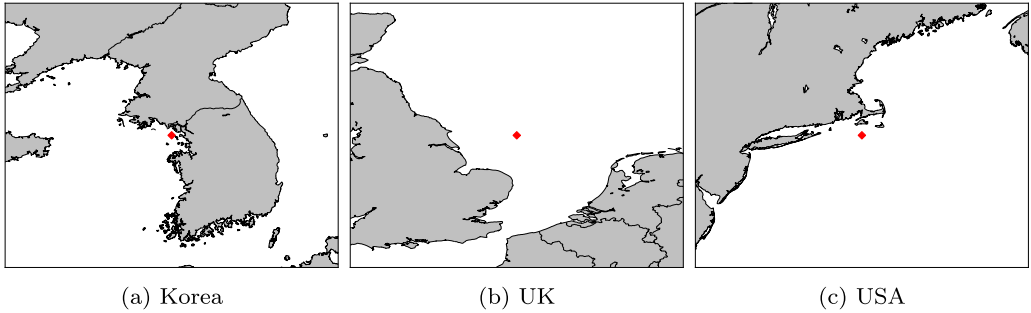}
    \caption{Maps with the three different prediction points and their surrounding area. The prediction points are represented by the red mark.}
    \label{fig:maps}
\end{figure}

The wind data are composed of the east-west wind speed ($u$) and north-south wind speed ($v$), at an altitude of 50 $m$. They cover the Korean Peninsula and surrounding areas, the UK and surrounding areas, and the northeastern USA, which allowed for a comparative analysis of the prediction performance across different regions (see Figure~\ref{fig:maps}). Moreover, the study focused on predicting wind power generation at individual points corresponding to real-world wind farms in the UK and USA, as well as one candidate site for a future wind farm in Korea. The spatial information of each dataset and the prediction points can be found in Table~\ref{tab:loc_tab}.

\begin{table}
\centering
\begin{adjustbox}{width=1\textwidth,center=\textwidth}
\small
\begin{tabular}{lrrrrrr}
\hline
  \textbf{Location}&
  \textbf{Lat. UL} &
  \textbf{Lat. LL} &
  \textbf{Lon. UL} &
  \textbf{Lon. LL} &
  \textbf{Prediction point} & \\

  \hline
    Korea  & 54.0$\degree$ N   & 20.5$\degree$ N  & 149.4$\degree$ E   & 105.0$\degree$ E  & 37.5$\degree$ N 126.3$\degree$ E \\
    UK   & 68.5$\degree$ N & 39.5$\degree$ N & 16.3$\degree$ E  & 21.9$\degree$ W & 54.0$\degree$ N 1.9$\degree$ E \\
    USA & 49.0$\degree$ N & 35.0$\degree$ N & 63.8$\degree$ W & 79.4$\degree$ W & 41$\degree$ N 70.6$\degree$ W \\ 
  \hline
\end{tabular}
\end{adjustbox}
\caption{Spatial information of the datasets. UL and LL denote the upper and lower limits, respectively.}
\label{tab:loc_tab}
\end{table}

The 3D-CNN was trained and tested using a 10-year period of data from \texttt{January 1, 2012} to \texttt{January 1, 2022}. The dataset was partitioned into three subsets: 60\% of the data were used for training (from \texttt{January 1, 2012} to \texttt{January 1, 2018}), 20\% for validation (from \texttt{January 2, 2018} to \texttt{January 1, 2020}), and the remaining 20\% for testing (from \texttt{January 2, 2020} to \texttt{January 1, 2022}).

\begin{figure}
    \centering
    \includegraphics[width = 1 \linewidth]{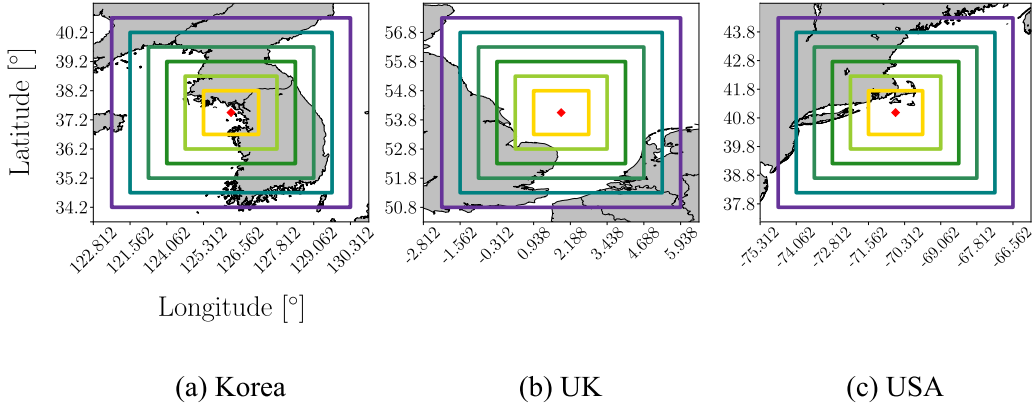}
    \caption{Visualization of the grid sizes in the three different regions.}
    \label{fig:grid_map}
\end{figure}

The impact of the surrounding information on wind prediction was investigated by incrementally increasing the latitude and longitude by $\Delta_{latitude}$ = 0.5\degree and $\Delta_{longitude}$ = 0.625\degree, respectively. The examined area began with a 3$\times$3 grid and gradually increased to a 13$\times$13 grid (refer to Figure~\ref{fig:grid_map}).

In the experiments, the effects of different time periods on wind flow prediction were studied. Specifically, time lengths of $T=3, 6, 12, 24$ h were considered. The CNN model used past wind data with a time interval of one hour to predict the wind flow for the upcoming hour. For example, when $T=6$, six consecutive snapshots of wind data obtained between \texttt{January 1, 2022, 06:00} and \texttt{January 1, 2022, 11:00} were used to forecast $u$ and $v$ values at \texttt{January 1, 2022, 12:00}.

\section{Methodology}
\label{method}

The methodology section explains the details of the employed (1) Atmospheric boundary layer (ABL) calibration and (2) Neural network model. The first subsection focuses on the ABL calibration method, which is employed to adapt the original wind data to the height of typical wind turbines. The architecture and hyperparameters of the employed 2D+3D-Convolutional Neural Network (2D+3D-CNN) model are explained in the second subsection.

\subsection{Atmospheric boundary layer calibration}
\label{sec:abl}
The wind data available from the MERRA-2 dataset were collected at a height of 50m. However, it is possible to approximate the wind velocity at different heights from the MERRA-2 dataset using the ABL calibration. This is particularly helpful when studying wind at heights where wind turbines are typically installed (e.g., 100m).

The ABL estimates wind speed and direction at different heights by accounting for the effects of atmospheric stability and turbulence on the wind profile~\cite{blocken2007cfd}. By the ABL method suggested by~\citet{richards1993appropriate}, the flow velocity $\vec{u} = (u,v)$ 
is calibrated to $\vec{u}_{ABL} = (u_{ABL},v_{ABL})$ by

\begin{linenomath}
    \begin{equation}
        \vec{u}_{ABL}(y) = \vec{u}_{ref}\frac{\ln(\frac{y_{ref}+y_0}{y_0})}{\ln(\frac{y+y_0}{y_0})},
    \label{eq:abl}
    \end{equation}
\end{linenomath}
where $y$ is the height of the wind field to be converted, $y_0=0.0002$ is the aerodynamic roughness length at sea level, $\vec{u}_{ABL}^*$ is the ABL friction velocity, $y_{ref}$ is set to $50m$, and $\vec{u}_{ref}$ is the wind velocity at an altitude of $50m$.
Using the ABL calibration, prediction and training of wind velocity at an altitude of $100m$ is demonstrated in this study.

After performing ABL calibration, the calibrated wind velocity ($u_{ABL}$ and $v_{ABL}$) is further standardized as
\begin{linenomath}
\begin{equation}
    u_{std} = \frac{u_{ABL}-\mu_{u_{ABL}}}{\sigma_{u_{ABL}}}
\label{eq:standard}
\end{equation}
\end{linenomath}
and
\begin{linenomath}
\begin{equation}
    v_{std} = \frac{v_{ABL}-\mu_{v_{ABL}}}{\sigma_{v_{ABL}}}
\label{eq:standard}
\end{equation}
\end{linenomath}
where $u_{std}$ and $v_{std}$ are the standardized velocity components, $\mu_{u_{ABL}}$ and $\mu_{v_{ABL}}$ are the mean of the velocity components, and $\sigma_{u_{ABL}}$ and $\sigma_{v_{ABL}}$ are the standard deviations of the velocity components. This standardization ensures that all features have similar scales and distributions.

\subsection{Neural network model}

\label{sec:ann}
\begin{figure}
    \centering
    \includegraphics[width = 1\linewidth]{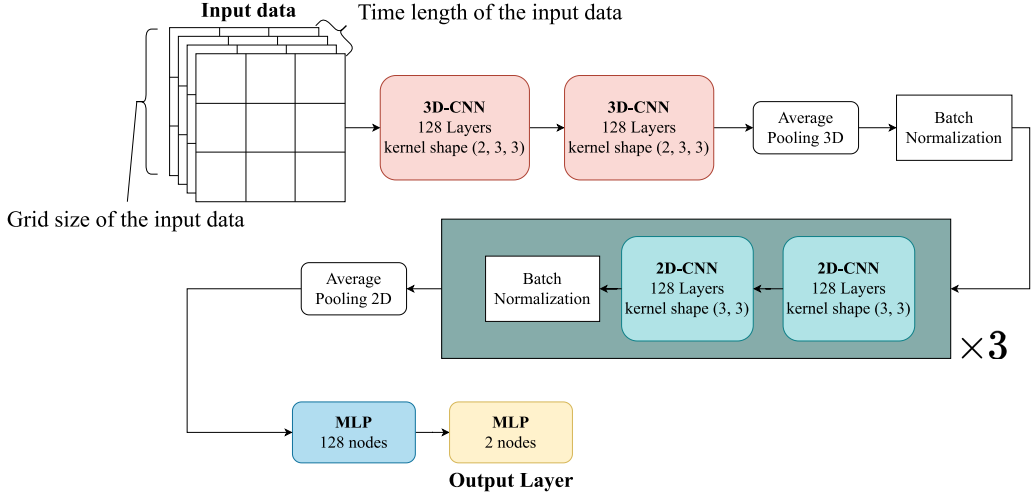}
    \caption{Schematic of the 3D-CNN model used in this study.}
    \label{fig:our_model}
\end{figure}

Neural networks have shown promising results in solving nonlinear problems and have been increasingly used in wind power research~\cite{shin2022neural, shahid2021novel, meka2021robust, sierra2021deep, jia2021reinforcement, howland2022collective}. In this study, a 2D+3D-CNN architecture based on the model proposed by \citet{higashiyama2018feature} was employed. The CNN is designed to learn spatiotemporal hierarchies of features in data by adjusting the weights and biases of convolutional filters. 3D convolutional filters move in three directions: 2D in space and 1D in time. A schematic of the 2D+3D-CNN model used in this paper can be seen in Figure\ref{fig:our_model}. This model employs the He uniform variance scaling initializer~\cite{he2015delving} for weights and biases of 2D and 3D convolutional filters, which is commonly used to facilitate more effective training of neural networks with Rectified Linear Unit (ReLU)-type activation functions. The range of the initial weights and biases is defined as

\begin{linenomath}
\begin{equation}
    (-\sqrt{\frac{6}{n_{in}}},\quad \sqrt{\frac{6}{n_{in}}}),
\label{eq:he}
\end{equation}
\end{linenomath}
$n_{in}$ represents the number of feature maps or nodes in the layer to be initialized. The leaky-ReLU activation function, a variation of the ReLU activation function~\cite{krizhevsky2017imagenet}, was employed. Leaky-ReLU is defined as
\begin{linenomath}
\begin{equation}
    f(x) = \max(\alpha x,x),
\label{eq:relu}
\end{equation}
\end{linenomath}
where $x$ is an arbitrary tensor and $\alpha = 0.3$ was used in this study. 

The Adaptive Moment Estimation (Adam) optimizer~\cite{kingma2014adam}, known for its efficient computational properties, its adaptive learning rate per parameter, and its use of both the first and second moments of the gradient, was used to train the model. 
Also, the model is trained to minimize the Huber loss~\cite{huber1992robust}.
This loss function is less sensitive to the presence of outliers in the training data, making it a more reliable measure of a model's performance in real-world scenarios. The Huber loss function is commonly employed in regression problems because it offers a balance between the mean squared error (MSE) and mean absolute error (MAE) loss functions, enabling it to handle both small and large deviations between the predicted and ground-truth values.
In addition, batch normalization layers are used to reduce overfitting and increase learning stability by shifting the layer inputs to zero mean and unit variance~\cite{ioffe2015batch}.
All neural network models used in this paper were implemented using \texttt{Python 3.10.6} and \texttt{Keras}~\cite{chollet2015keras}. Additionally, the models were trained using two \texttt{NVIDIA GeForce RTX 3060} graphics processing units (GPU).

\section{Results}
\label{results}
The coefficient of determination $R^2$ is used to evaluate the prediction performance and is defined as 

\begin{linenomath}
\begin{equation}
    R^2 = \frac{\sum\limits_{i=1}^n(\hat{y_i}-\bar{y})^2}{\sum\limits_{i=1}^n(y_i - \bar{y})^2},
\label{eq:r2}
\end{equation}
\end{linenomath}
where $y_i$ is the ground truth value, $\bar{y}$ is the mean of the ground truth values and $\hat{y_i}$ is the predicted value. The value of $R^2$ ranges from 0 to 1, with a value closer to 1 indicating a more accurate model and a value closer to 0 indicating a less accurate model. 

\begin{figure}
  \centering
        \includegraphics[width=\linewidth]{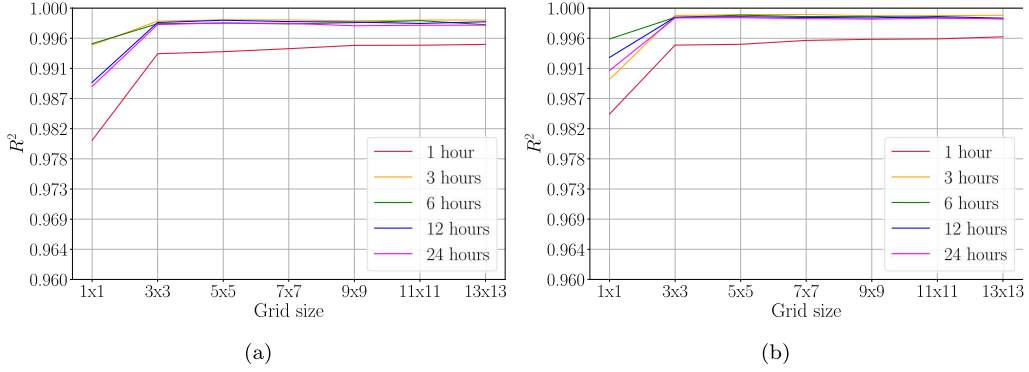}
    \caption{Variations in $R^2$ values for predicting $u$ (a) and $v$ (b) with changing grid sizes and input time lengths in Korea.}
    \label{fig:korea_r2}
\end{figure}

\begin{table}[h!]
    \begin{minipage}{.9\textwidth}
        \centering
        \begin{tabular}{|c|c|c|c|c|c|c|c|c|}
            \hline
             $u$ &\textbf{1$\times$1} & \textbf{3$\times$3} & \textbf{5$\times$5} & \textbf{7$\times$7} & \textbf{9$\times$9} & \textbf{11$\times$11} & \textbf{13$\times$13}\\ \hline
            \textbf{1} & 0.970 & 0.983 & 0.986 & 0.987 & 0.986 & 0.986 & 0.986 \\ \hline
            \textbf{3} & 0.992 & 0.995 & 0.996 & 0.996 & 0.996 & 0.995 & 0.996 \\ \hline
            \textbf{6} & 0.989 & 0.995 & 0.996 & 0.996 & 0.995 & 0.994 & 0.995 \\ \hline
            \textbf{12} & 0.985 & 0.994 & 0.994 & 0.994 & 0.993 & 0.993 & 0.992 \\ \hline
            \textbf{24} & 0.982 & 0.993 & 0.993 & 0.993 & 0.992 & 0.992 & 0.992 \\ \hline
        \end{tabular}
        \\
        \vspace{0.5cm}

        \begin{tabular}{|c|c|c|c|c|c|c|c|c|}
            \hline
            $v$ &\textbf{1$\times$1} & \textbf{3$\times$3} & \textbf{5$\times$5} & \textbf{7$\times$7} & \textbf{9$\times$9} & \textbf{11$\times$11} & \textbf{13$\times$13}\\ \hline
            \textbf{1} & 0.968 & 0.987 & 0.988 & 0.989 & 0.989 & 0.988 & 0.989 \\ \hline
            \textbf{3} & 0.992 & 0.996 & 0.997 & 0.996 & 0.996 & 0.996 & 0.996 \\ \hline
            \textbf{6} & 0.989 & 0.996 & 0.996 & 0.996 & 0.995 & 0.995 & 0.995 \\ \hline
            \textbf{12} & 0.985 & 0.995 & 0.994 & 0.995 & 0.994 & 0.993 & 0.993 \\ \hline
            \textbf{24} & 0.982 & 0.994 & 0.994 & 0.994 & 0.993 & 0.993 & 0.992 \\ \hline
        \end{tabular}  
        \caption{$R^2$ values for predicting $u$, $v$ with changing grid sizes and input time lengths in Korea.}
    \label{tab:r2_korea}
    \end{minipage}
\end{table}

\begin{figure}
  \centering
        \includegraphics[width=\linewidth]{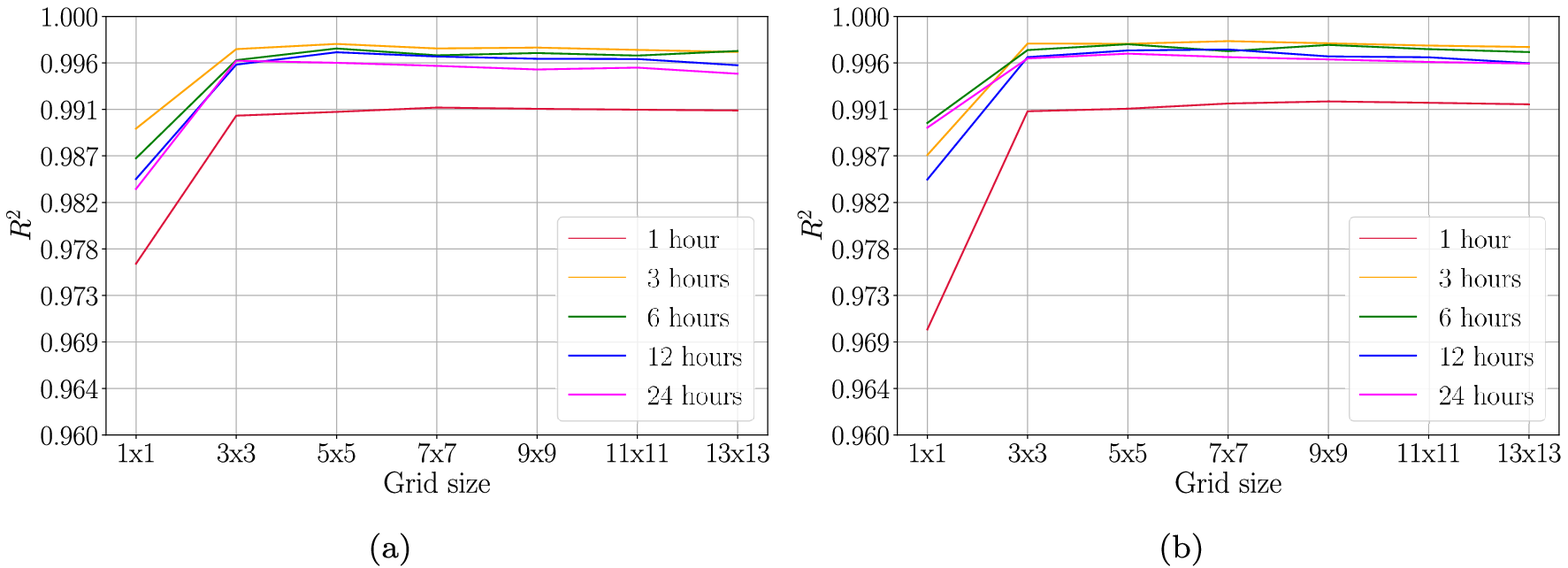}
    \caption{Variations in $R^2$ values for predicting $u$ (a) and $v$ (b) with changing grid sizes and input time lengths in the USA.}
    \label{fig:USA_r2}
\end{figure}

\begin{table}[h!]
    \begin{minipage}{.9\textwidth}
    \centering
    \begin{tabular}{|c|c|c|c|c|c|c|c|c|}
        \hline
         $u$ &\textbf{1$\times$1} & \textbf{3$\times$3} & \textbf{5$\times$5} & \textbf{7$\times$7} & \textbf{9$\times$9} & \textbf{11$\times$11} & \textbf{13$\times$13}\\ \hline
        \textbf{1} & 0.976 & 0.991 & 0.991 & 0.991 & 0.991 & 0.991 & 0.991 \\ \hline
        \textbf{3} & 0.989 & 0.997 & 0.997 & 0.997 & 0.997 & 0.997 & 0.997 \\ \hline
        \textbf{6} & 0.986 & 0.996 & 0.997 & 0.996 & 0.996 & 0.996 & 0.997 \\ \hline
        \textbf{12} & 0.984 & 0.995 & 0.997 & 0.996 & 0.996 & 0.996 & 0.995 \\ \hline
        \textbf{24} & 0.984 & 0.996 & 0.996 & 0.995 & 0.995 & 0.995 & 0.995 \\ \hline
    \end{tabular}
    \\
    \vspace{0.5cm}
    \begin{tabular}{|c|c|c|c|c|c|c|c|c|}
        \hline
         $v$ &\textbf{1$\times$1} & \textbf{3$\times$3} & \textbf{5$\times$5} & \textbf{7$\times$7} & \textbf{9$\times$9} & \textbf{11$\times$11} & \textbf{13$\times$13}\\ \hline
        \textbf{1} & 0.970 & 0.991 & 0.991 & 0.992 & 0.992 & 0.992 & 0.992  \\ \hline
        \textbf{3} & 0.987 & 0.997 & 0.997 & 0.998 & 0.997 & 0.997 & 0.997  \\ \hline
        \textbf{6} & 0.990 & 0.997 & 0.997 & 0.997 & 0.997 & 0.997 & 0.997  \\ \hline
        \textbf{12} & 0.984 & 0.996 & 0.997 & 0.997 & 0.996 & 0.996 & 0.996  \\ \hline
        \textbf{24} & 0.989 & 0.996 & 0.996 & 0.996 & 0.996 & 0.996 & 0.995  \\ \hline
    \end{tabular}
    \caption{$R^2$ values for predicting $u$, $v$ with changing grid sizes and input time lengths in the USA.}
    \label{tab:r2_USA}
    \end{minipage}
\end{table}

\begin{figure}
  \centering
        \includegraphics[width=\linewidth]{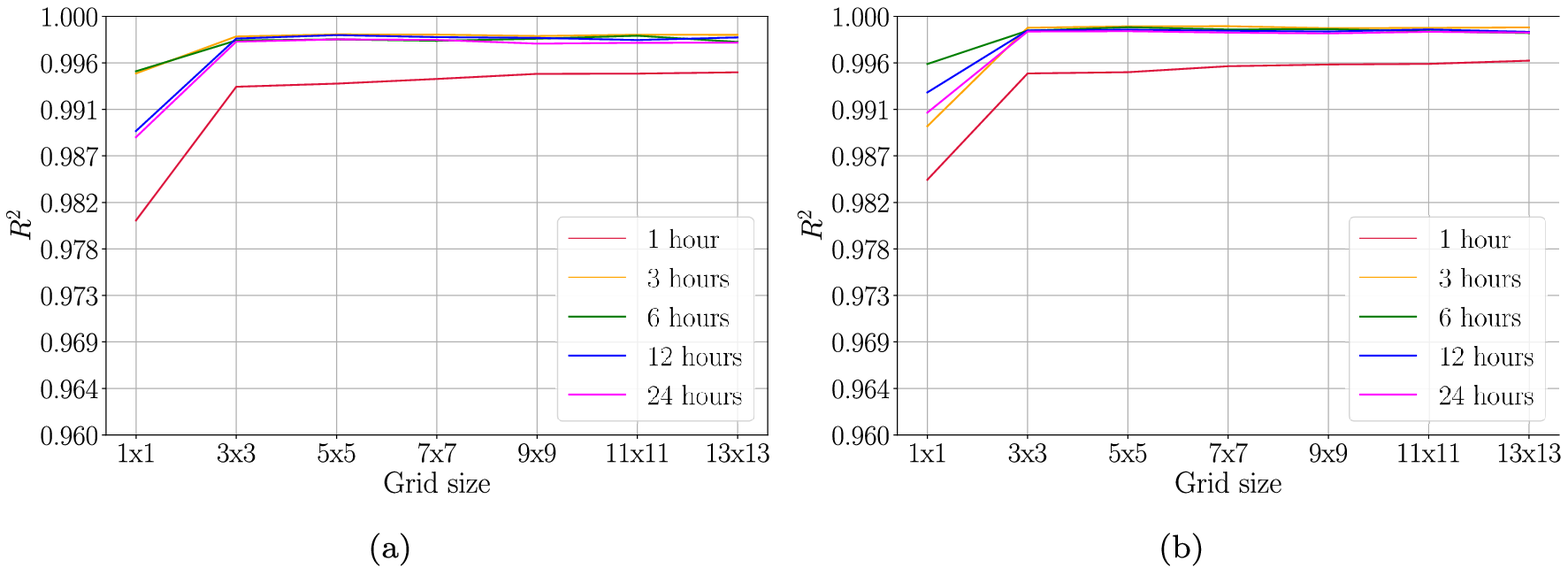}
    \caption{Variations in $R^2$ values for predicting $u$ (a) and $v$ (b) with changing grid sizes and input time lengths in the UK.}
    \label{fig:UK_r2}
\end{figure}

\begin{table}[h!]
    \begin{minipage}{.9\textwidth}
        \centering
        \begin{tabular}{|c|c|c|c|c|c|c|c|c|}
            \hline
             $u$ &\textbf{1$\times$1} & \textbf{3$\times$3} & \textbf{5$\times$5} & \textbf{7$\times$7} & \textbf{9$\times$9} & \textbf{11$\times$11} & \textbf{13$\times$13}\\ \hline
            \textbf{1} & 0.981 & 0.993 & 0.994 & 0.994 & 0.995 & 0.995 & 0.995   \\ \hline
            \textbf{3} & 0.995 & 0.998 & 0.998 & 0.998 & 0.998 & 0.998 & 0.998   \\ \hline
            \textbf{6} & 0.995 & 0.998 & 0.998 & 0.998 & 0.998 & 0.998 & 0.998   \\ \hline
            \textbf{12} & 0.989 & 0.998 & 0.998 & 0.998 & 0.998 & 0.998 & 0.998  \\ \hline
            \textbf{24} & 0.988 & 0.998 & 0.998 & 0.998 & 0.997 & 0.997 & 0.997  \\ \hline
            \end{tabular}
        \\
        \vspace{0.5cm}
        \begin{tabular}{|c|c|c|c|c|c|c|c|c|}
            \hline
             $v$ &\textbf{1$\times$1} & \textbf{3$\times$3} & \textbf{5$\times$5} & \textbf{7$\times$7} & \textbf{9$\times$9} & \textbf{11$\times$11} & \textbf{13$\times$13}\\ \hline
            \textbf{1} & 0.984 & 0.995 & 0.995 & 0.995 & 0.995 & 0.995 & 0.996   \\ \hline
            \textbf{3} & 0.990 & 0.999 & 0.999 & 0.999 & 0.999 & 0.999 & 0.999  \\ \hline
            \textbf{6} & 0.995 & 0.999 & 0.999 & 0.999 & 0.999 & 0.999 & 0.998   \\ \hline
            \textbf{12} & 0.993 & 0.999 & 0.999 & 0.999 & 0.999 & 0.999 & 0.999   \\ \hline
            \textbf{24} & 0.991 & 0.999 & 0.999 & 0.998 & 0.998 & 0.999 & 0.998  \\ \hline
    \end{tabular}
    \caption{$R^2$ values for predicting $u$, $v$ with changing grid sizes and input time lengths in the UK.}
    \label{tab:r2_UK}
    \end{minipage}
\end{table}

The evaluation of the prediction performance with respect to the change in the spatiotemporal size of the input data is presented in Figures~\ref{fig:korea_r2}–\ref{fig:UK_r2} and Tables~\ref{tab:r2_korea}-\ref{tab:r2_UK}.
Each color represents a different time length for the input data.
In terms of spatial aspects, providing additional surrounding data to the 2D+3D-CNN model improved the prediction accuracy compared to providing a single space ($1\times1$), regardless of the time length.
However, no significant difference was observed in the values of $R^2$ when the spatial area was increased by more than $3\times3$. 

Similarly, in terms of the time length, using a single time step resulted in the worst prediction performance in every case.
The lower accuracy achieved with a single-time-step input can be attributed to the inherent limitations in capturing temporal trends with only a single snapshot.
Overall, the best predictions were obtained at a time of $3$ hours.
No significant difference was observed in the values of $R^2$ between the 3 h and 24 h time periods.

\begin{figure}
  \centering
        \includegraphics[width=\linewidth]{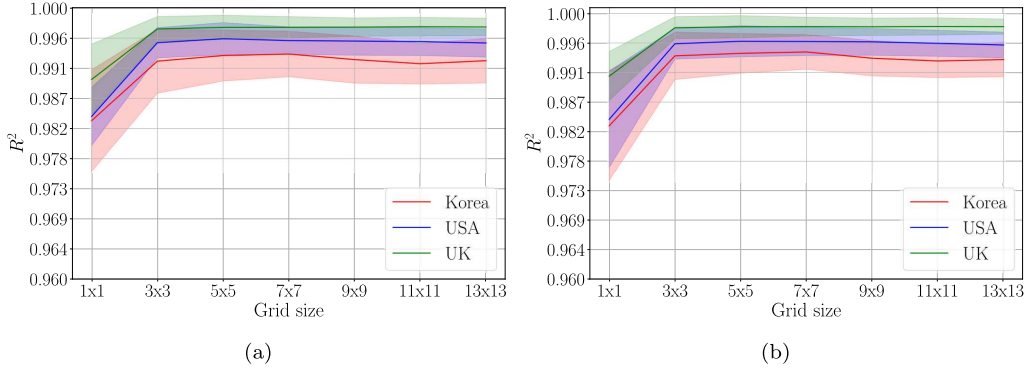}
    \caption{Comparison of $R^2$ with varying spatial input data for predicting $u$ (a) and $v$ (b). The solid lines represent the average $R^2$ values across all time lengths, and the shaded area indicates the range of minimum and maximum $R^2$ values.}
    \label{fig:mc_entire_3d}
\end{figure}

In contrast, the impact of regional differences on observed variations is noteworthy. Figure~\ref{fig:mc_entire_3d} provides a clear visualization of the regional discrepancies in predicting $u$ and $v$. Korea has the highest variance in $R^2$ compared to the USA and UK. In addition, Korea has the lowest average $R^2$ value for all grid sizes.

To investigate the cause of the performance differences resulting from the different temporal flow scales in the tested regions, an autocorrelation analysis was employed. Autocorrelation measures the linear relationship between time-series data and their shifted versions. A low autocorrelation coefficient (ACC) indicates a weak correlation between the original data and the same data shifted by a certain time lag, whereas a high ACC suggests a strong correlation between the original and shifted data. ACC can be calculated as

\begin{linenomath}
\begin{equation}
    r_k = \frac{\sum\limits_{t=k+1}^{n}(y_t-\bar{y})(y_{t-k}-\bar{y})}{\sum\limits_{t=1}^{n}(y_t-\bar{y})^2},
\label{eq:acc}
\end{equation}
\end{linenomath}
where $r_k$ represents the ACC at lag $k$, $y_t$ is the value of the time series at time $t$, and $n$ is the total number of time steps. $\bar{y}$ represents the mean of the time series.

\begin{figure}
    \centering
            \includegraphics[width=\linewidth]{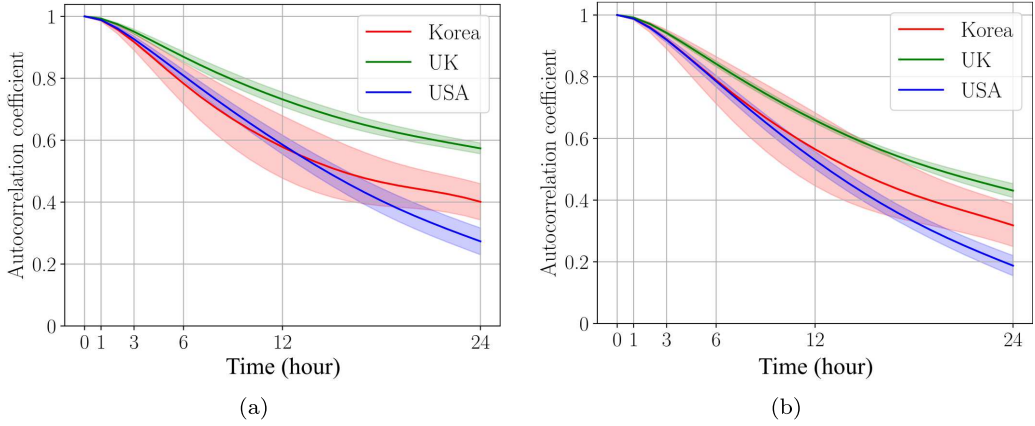}
    \caption{ACCs of (a) $u$  and (b) $v$. The y-axis represents the ACC values, while the x-axis represents the input time lengths. The solid line indicates the mean value, and the shaded area stands for the minimum and maximum ACC values.}
    \label{fig:ACF}
\end{figure}

\begin{figure}
    \centering
            \includegraphics[width=\linewidth]{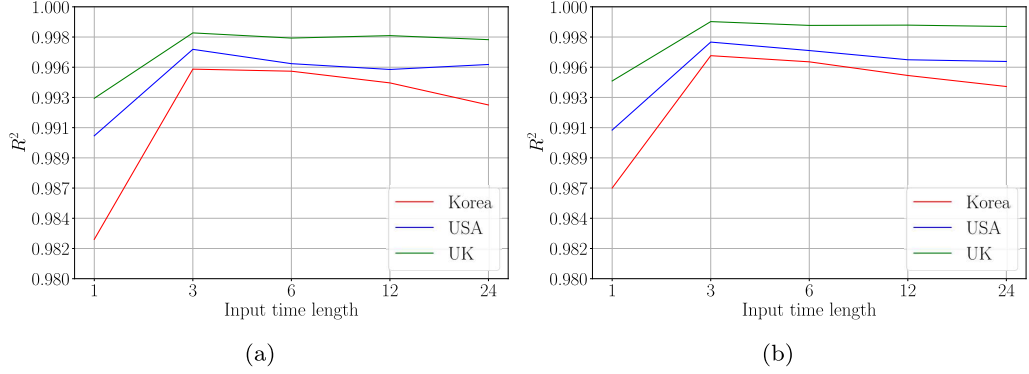}
    \caption{Effect of increasing input time length on performance ($R^{2}$) with a fixed grid size of $3\times3$ for $u$ (a) and $v$ (b).}
    \label{fig:TC}
\end{figure}

Figure~\ref{fig:ACF} shows the ACC values for the $13\times13$ grid space. As shown in the figure, the ACC values for all three regions steadily decreased as the time lag increased. However, the rate of decrease varied across regions, with the ACC of the UK displaying the gentlest slope, whereas those of Korea and the USA exhibited steeper slopes. This trend aligns with the observations in Figure~\ref{fig:TC} and Table~\ref{tab:r2_3by3}, which highlight a larger change in $R^2$ with changing input time lengths in Korea and the USA compared to the UK.

Furthermore, it is worth noting that the shaded areas in Figure~\ref{fig:ACF} vary across different regions. These shaded areas are represented based on the standard deviation calculated as
\begin{linenomath}
\begin{equation}
    \sigma_{k} = \sqrt{\frac{{\sum_{i=1}^{n}(r_{k,i} - \mu_{k})^2}}{n}},
\label{eq:acc}
\end{equation}
\end{linenomath}
where $\sigma_{k}$ is the standard deviation of ACC values at time lag $k$, $r_{k,i}$ is the ACC value at time lag $k$ and $i$-th location in the grid space, $\mu_{k}$ is the spatial average of $r_{k,i}$, and $n$ is the total number of spatial grid points.
Furthermore, the maximum standard deviations across different regions over time were calculated as $\max_{k}(\sigma_{k})$. These results are presented in Table~\ref{tab:max_std}.
The regions with the highest maximum standard deviations were found to be in the following order: Korea, USA, and UK; while the predictive performances ($R^{2}$) were observed to be the best in the reverse order: UK, USA, and Korea. This result indicates that a CNN's learnability is highly dependent on regional spatiotemporal wind characteristics that could be quantified by calculating ACCs.

\begin{table}[]
\centering
\begin{tabular}{|c|l|l|l|}
\hline
\multicolumn{1}{|l|}{} & \multicolumn{1}{c|}{\textbf{Korea}} & \multicolumn{1}{c|}{\textbf{USA}} & \multicolumn{1}{c|}{\textbf{UK}} \\ \hline
\textbf{$u$}          & 0.101                            & 0.043                         & 0.023                         \\ \hline
\textbf{$v$}          & 0.119                            & 0.033                         & 0.022                         \\ \hline
\end{tabular}%
\caption{The maximum standard deviation of ACC of $u$, $v$ values for each region.}
\label{tab:max_std}
\end{table}

\begin{table}[h!]
\begin{minipage}{0.9 \textwidth}
    \centering
    \begin{tabular}{|c|c|c|c|c|c|c|c|c|}
        \hline
        $u$ &\textbf{1} & \textbf{3} & \textbf{6} & \textbf{12} & \textbf{24}\\ \hline
        \textbf{Korea} & 0.983 & 0.995 & 0.995 & 0.994 & 0.993 \\ \hline
        \textbf{USA} & 0.991 & 0.997 & 0.996 & 0.995 & 0.996 \\ \hline
        \textbf{UK} & 0.993 & 0.998 & 0.998 & 0.998 & 0.998 \\ \hline
    \end{tabular}
    \\
    \vspace{0.5cm}
    \centering
    \begin{tabular}{|c|c|c|c|c|c|c|c|c|}
        \hline
        $v$ &\textbf{1} & \textbf{3} & \textbf{6} & \textbf{12} & \textbf{24}\\ \hline
        \textbf{Korea} & 0.987 & 0.996 & 0.996 & 0.995 & 0.994 \\ \hline
        \textbf{USA} & 0.991 & 0.997 & 0.997 & 0.996 & 0.996 \\ \hline
        \textbf{UK} & 0.995 & 0.999 & 0.999 & 0.999 & 0.999 \\ \hline
        
    \end{tabular}
    \caption{Effect of increasing input time length on performance ($R^{2}$) with a fixed grid size of $3\times3$ for $u$, $v$}
    \label{tab:r2_3by3}
\end{minipage}
\end{table}

To further investigate the spatial effect, a Pearson correlation coefficient (PCC) analysis was conducted. PCC is a statistical tool used to quantify the linear correlation between variables. The PCC of the two variables, $a$ and $b$, is defined by

\begin{linenomath}
\begin{equation}
    PCC = \frac{\sum\limits_{i=1}^{n}(a_i-\bar{a})(b_i-\bar{b})}{\sqrt{\sum\limits_{i=1}^{n}(a_i-\bar{a})^2}\sqrt{\sum\limits_{i=1}^{n}(b_i-\bar{b})^2}},
\label{eq:pcc}
\end{equation}
\end{linenomath}
where $\bar{a}$ and $\bar{b}$ indicate the mean values of $a$ and $b$, and $n$ denotes the number of data points.

\begin{figure}
\begin{minipage}{0.95\textwidth}
    \centering
        \includegraphics[width=\linewidth]{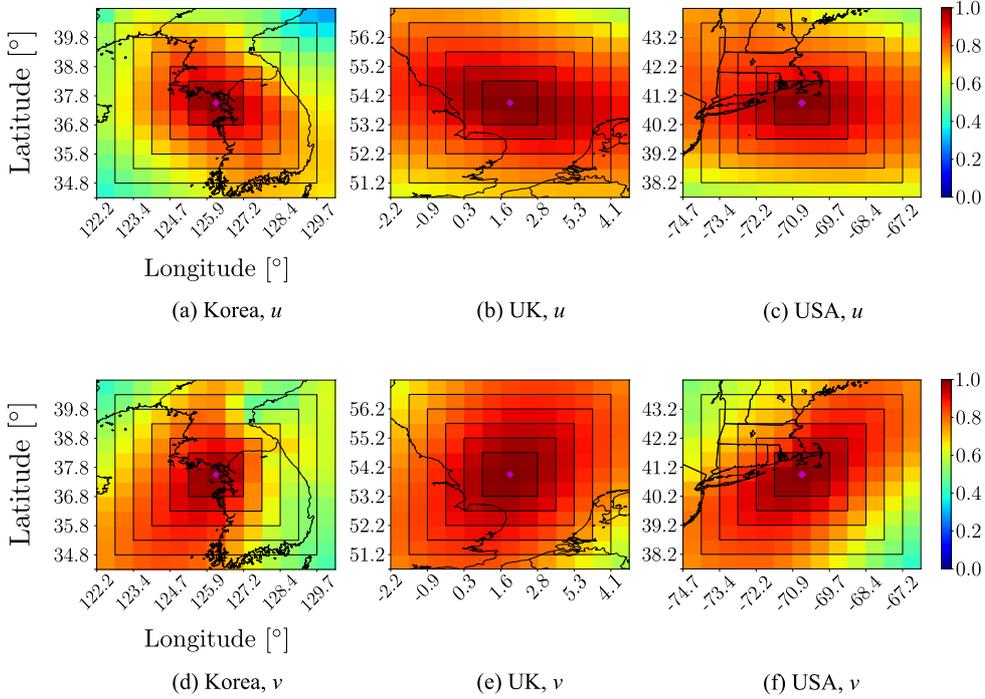}
    \caption{The PCC heatmaps of variable $u$ across three regions over a 10-year period are depicted in (a) to (c), while the PCC heatmaps of variable $v$ during the same period in these regions are shown in (d) to (f).}
    \label{fig:PCC}
    \end{minipage}
\end{figure}

\begin{figure}
    \centering
            \includegraphics[width=\linewidth]{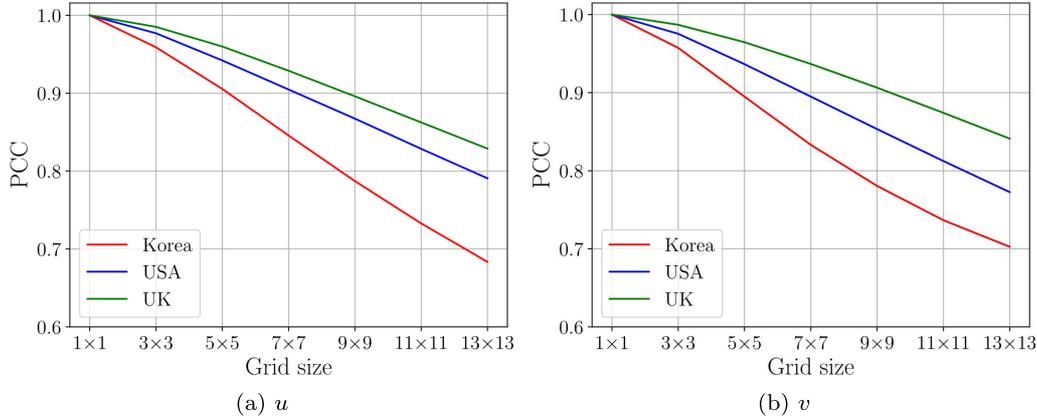}
    \caption{Changes in average PCC for (a) $u$ and (b) $v$ as grid size increases.}
    \label{fig:avg_PCC}
\end{figure}

The variation in the PCC with respect to spatial size was examined to determine the cause of the difference in forecasting performance by region. Ten years of wind data were used to compute the PCC of both $u$ and $v$ at each grid point and prediction point. The results are presented in the form of heatmaps (Figure~\ref{fig:PCC}). The heatmaps revealed that the PCC values for all three locations were similar up to a grid size of $3\times3$. For grid sizes of $5\times5$ and larger, the PCC values in Korea were lower than those in the UK and USA for both velocity components. Among the three regions, the UK exhibited the highest concentration of high PCC values. Figure~\ref{fig:avg_PCC} shows the average values of PCC with respect to grid size for each region, excluding the prediction point. The UK displayed the smallest change in the average $u$ and $v$ PCC.

\begin{figure}
    \centering
        \includegraphics[width=\linewidth]{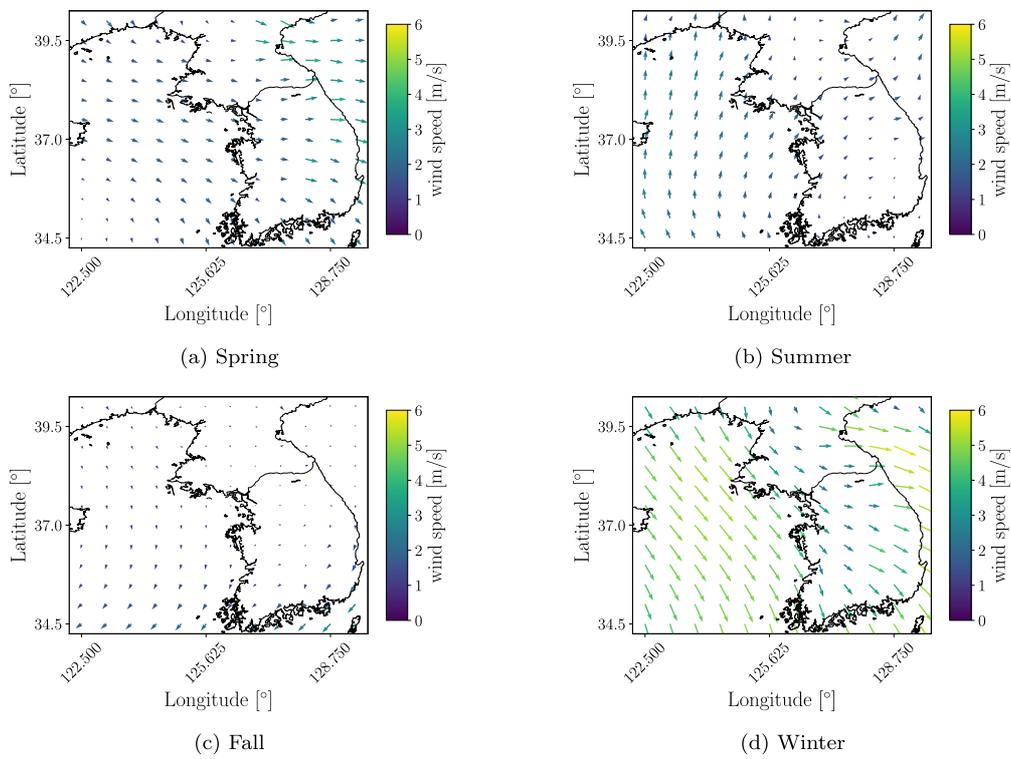}
    \caption{Wind field variations in Korea in four distinct seasons. The size of the arrow is proportional to the speed.}
    \label{fig:Korea_windfield}
\end{figure}

\begin{figure}
    \centering
        \includegraphics[width=\linewidth]{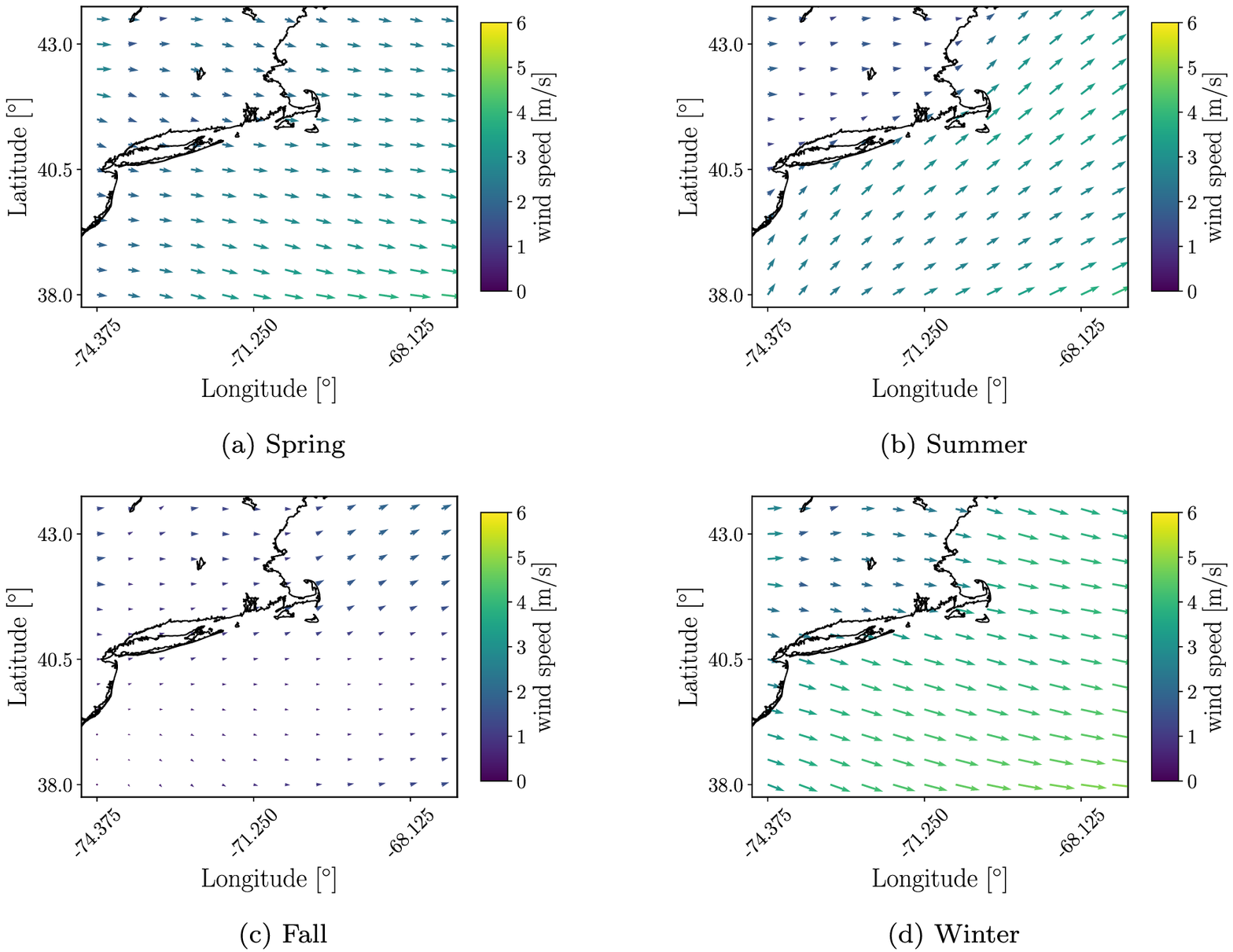}
    \caption{Wind field variations in the USA in four distinct seasons. The size of the arrow is proportional to the speed.}
    \label{fig:USA_windfield}
\end{figure}

\begin{figure}
    \centering
        \includegraphics[width=\linewidth]{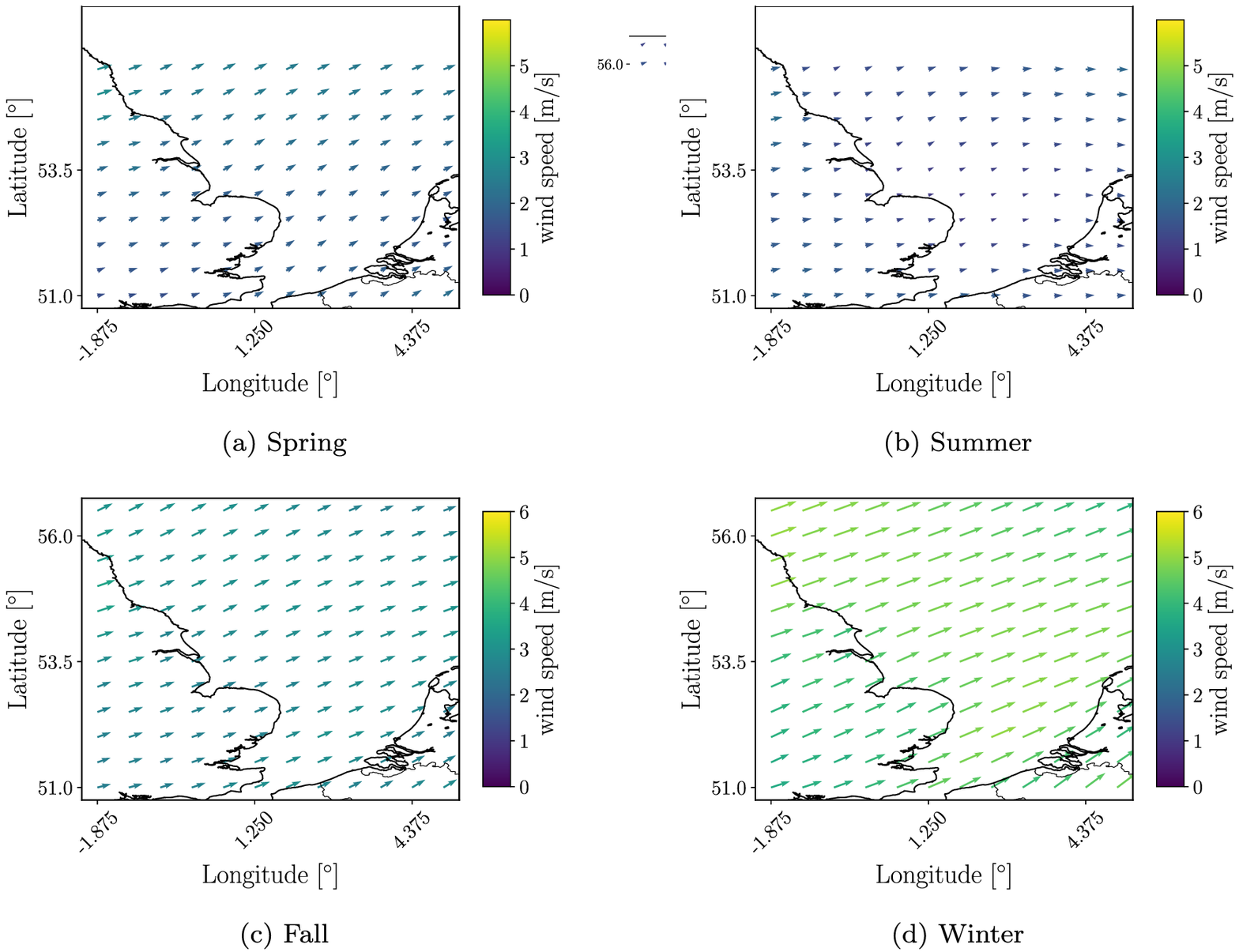}
    \caption{Wind field variations in the UK in four distinct seasons. The size of the arrow is proportional to the speed.}
    \label{fig:UK_windfield}
\end{figure}

The observed differences in PCC values across the three regions can be attributed to their respective seasonal wind patterns. A closer look at the wind fields in Korea, the USA, and the UK, as depicted in Figures~\ref{fig:Korea_windfield}, \ref{fig:USA_windfield}, and \ref{fig:UK_windfield}, highlights that Korea exhibits the most intricate and complex wind flow patterns unique to the region. During the winter, Korea is influenced by northwest monsoons from Siberia, as shown in Figure~\ref{fig:Korea_windfield} (d), whereas during the summer, it is influenced by southeast monsoons from the North Pacific, as shown in Figure~\ref{fig:Korea_windfield} (b).
Moreover, the complex terrain of Korea, with its many mountains, greatly affects the flow of the near-surface atmosphere, contributing to intricate wind patterns. In addition, Korea is affected by strong tropical cyclones, known as typhoons, during the summer and fall, further increasing the variability in wind patterns~\cite{kim2013spatio}.

In contrast, the PCC values in the UK and the USA showed a more even distribution with higher overall values. This can be attributed to the topographic and meteorological features of these regions. The UK, in particular, has a more gentle and flatter terrain than Korea, which does not significantly impede the wind flow in the region. The prevailing winds in this region are westerlies that blow from the Atlantic Ocean and are relatively consistent throughout the year.

Similarly, the prevailing winds in the northeastern region of the US blowing from the Pacific Ocean are westerlies. During the winter, the Northeastern USA can experience extreme weather conditions due to the influence of the "Polar Vortex"~\cite{overland2019impact}. However, compared to the wind patterns in Korea, the wind patterns in both the Northeastern UK and the USA are relatively consistent throughout the year, which is reflected by a more even distribution of PCC values in these regions.

A comparative analysis was conducted to compare the predictive capability of the 2D+3D-CNN model employed in this research against state-of-the-art wind prediction CNN models. \citet{zhu2021wind} introduced a Fully 3D-CNN model to capture spatiotemporal information, which is similar to the proposed 2D+3D-CNN model. The main distinction between the Fully 3D-CNN model and the current 2D+3D-CNN model is that the Fully 3D-CNN model does not include 2D-CNN layers.
\citet{shen2022wind} employed a 1D-CNN model consisting of long short-term memory (LSTM) layers~\cite{hochreiter1997long, greff2016lstm}, known for its ability to learn long-term dependencies in sequential data, to predict wind speeds for unmanned sailboat control systems.

The comparison was conducted using data with grid sizes ranging from $7\times7$ to $13\times13$ and input time lengths of 6, 12, and 24, due to structural constraints in the models of the aforementioned studies. In this study, the last fully connected layer of the Fully 3D-CNN model, which originally consisted of seven neurons, was modified to have two neurons for predicting $u$ and $v$ at future time steps. Similarly, the 1D-CNN was adjusted to include two cells in the last LSTM layer.

\begin{figure}
        \includegraphics[width=\linewidth]{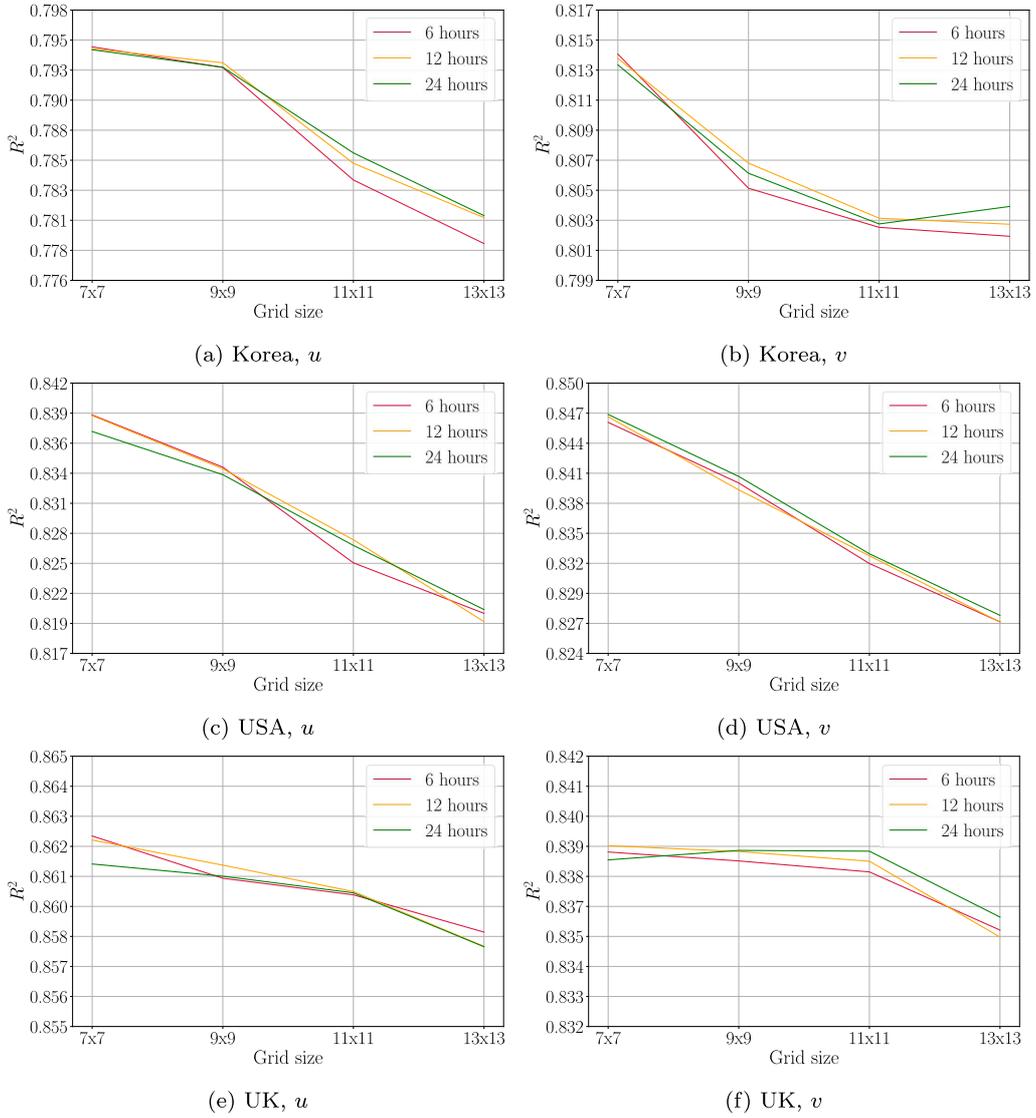}
    \caption{Variations in $R^2$ values for predicting $u$ and $v$ with changing grid sizes and input time lengths using the 1D-CNN model in Korea, USA, and UK.}
    \label{fig:r2_Shen}
\end{figure}

\begin{figure}
        \centering
        \includegraphics[width=\linewidth]{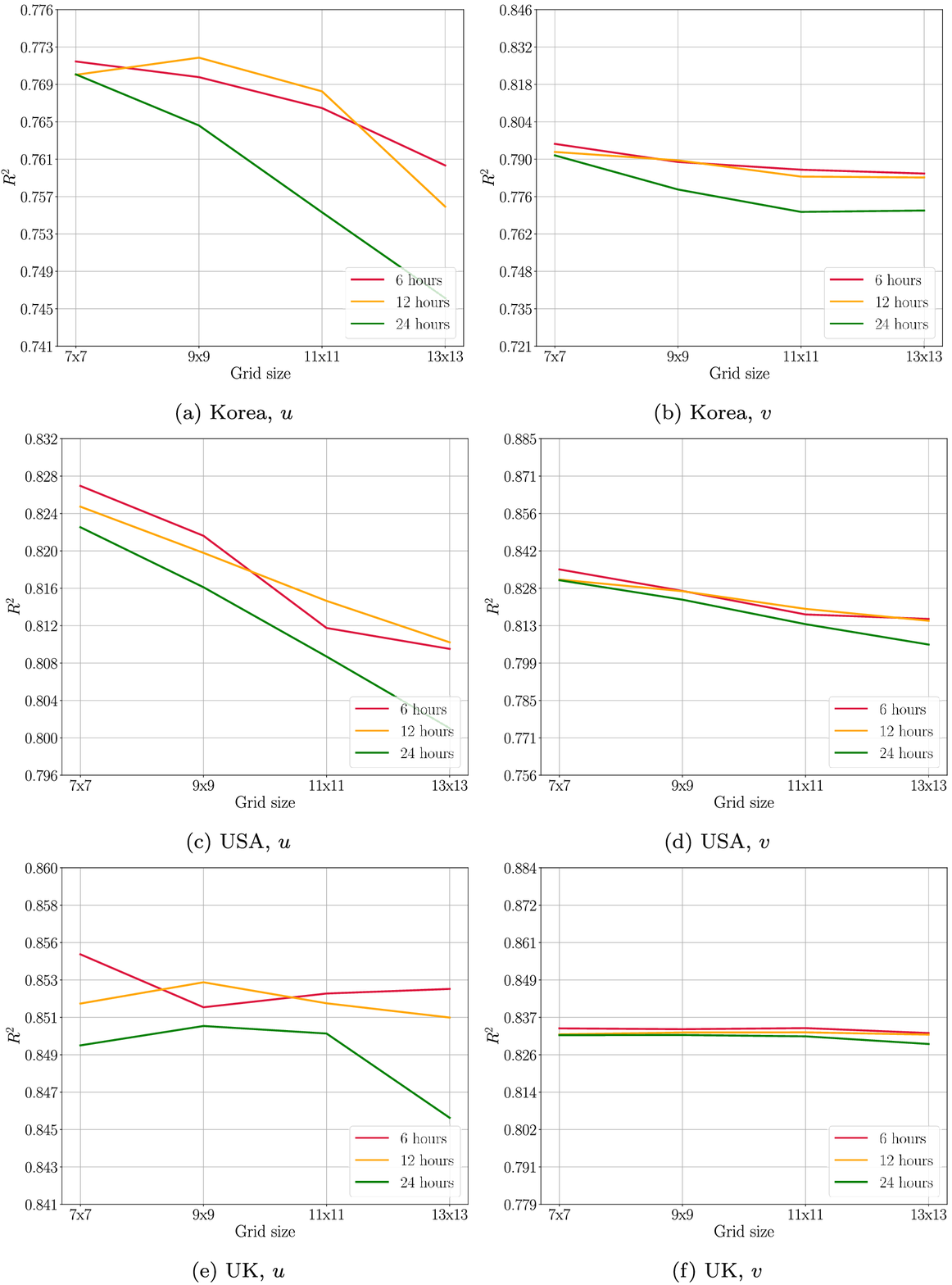}
    \caption{Variations in $R^2$ values for predicting $u$ and $v$ with changing grid sizes and input time lengths using the fully 3D-CNN model in Korea, USA, and UK.}
    \label{fig:r2_Zhu}
\end{figure}

The predictive performances ($R^{2}$) of the 1D-CNN and Fully 3D-CNN models were analyzed by varying the grid and time step sizes, as shown in Figures~\ref{fig:r2_Shen}~and~\ref{fig:r2_Zhu}, respectively. The 1D-CNN model demonstrated similar performance even when using a longer history of time for wind prediction. In addition, the performance decreased when larger spatial information was included for training. Similarly, the Fully 3D-CNN model tended to show a performance drop when larger spatial data was included in the input. This observation was also consistent with the findings from the 2D+3D-CNN model when data with smaller spatiotemporal correlations (i.e., spatial data larger than $7 \times 7$) were provided (Figure~\ref{fig:mc_entire_3d}).

\begin{figure}
    \centering
            \includegraphics[width=\linewidth]{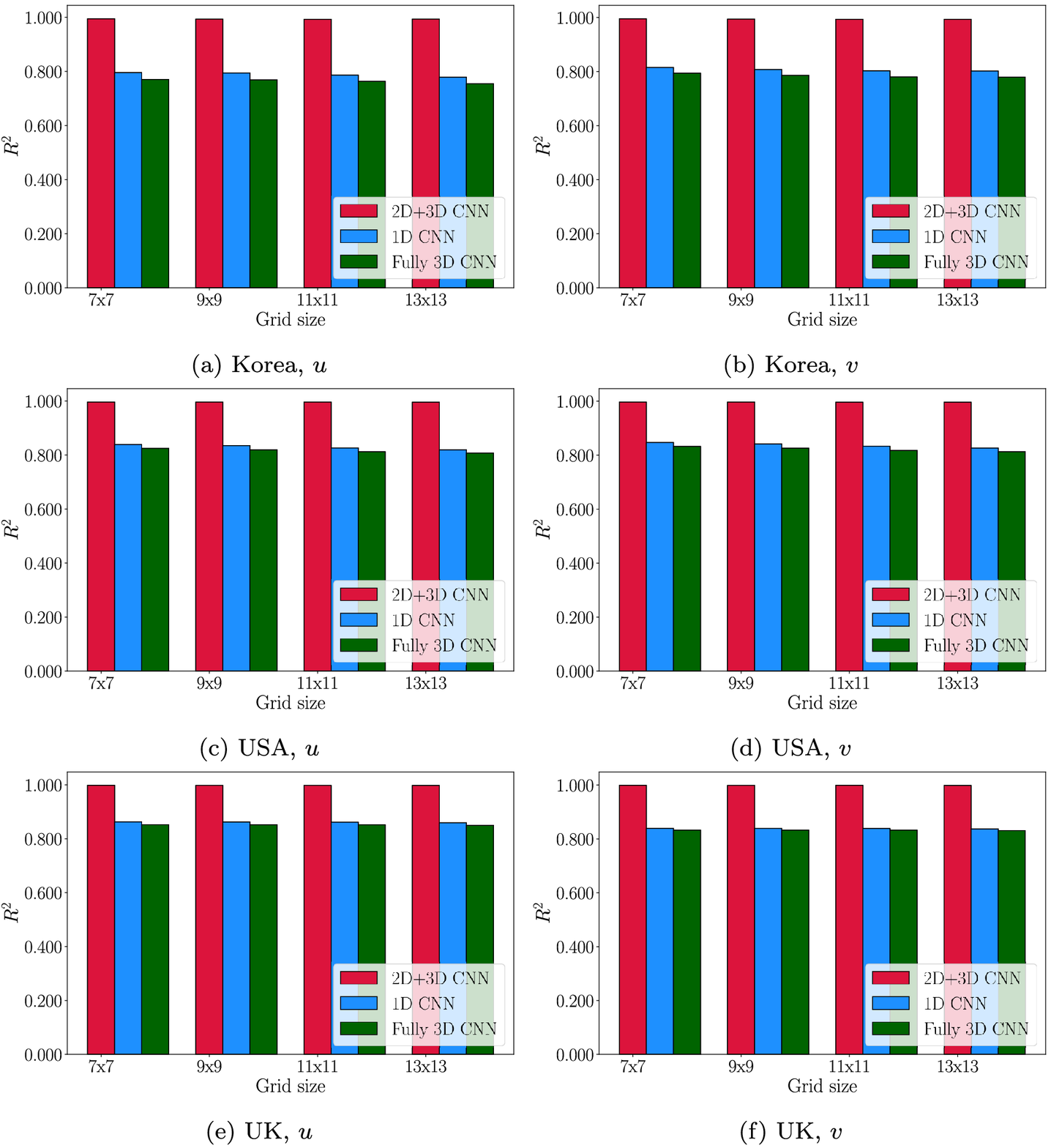}
    \caption{Comparison of average $R^2$ in predicting (a) $u$ and (b) $v$ between models by grid size in Korea, USA, and UK. The models by \citet{zhu2021wind}, \citet{shen2022wind}, and the current study are represented in the figure as Fully 3D CNN, 1D CNN, and 2D+3D CNN, respectively.}
    \label{fig:model_compare}
\end{figure}

Overall, the present 2D+3D-CNN model demonstrates a significant performance improvement compared to the Fully 3D-CNN and 1D-CNN models (Figure~\ref{fig:model_compare}). This improvement can be attributed to the deep layers employed in the present 2D+3D-CNN model, which allows for efficient spatiotemporal feature extraction by utilizing both 2D and 3D convolutional layers. However, it is important to note that the 2D+3D-CNN model is specifically designed for spatiotemporal data with uniform resolutions in both space and time, whereas the 1D-CNN and Fully 3D-CNN models were developed using real-world measurement data that are spatially sparse~\cite{zhu2021wind, shen2022wind}. As a result, the superior performance of the current model may not be guaranteed when applied to spatially sparse data.

Nevertheless, it is worth noting that all of the tested CNN models exhibit the best predictive performance in the order of regions in the UK, the USA, and Korea (Figures~\ref{fig:mc_entire_3d},~\ref{fig:r2_Shen}~,~\ref{fig:r2_Zhu},~and~\ref{fig:model_compare}). This result suggests that regional wind characteristics have a substantial impact on the performance of the predictive CNN models. Moreover, the regional wind effects on the neural network's learnability can be estimated prior to training through the spatiotemporal correlation analysis proposed in this study.

\section{Conclusions}
This paper investigates how the performance of wind prediction CNN models can be influenced by regional wind characteristics. The proposed 2D+3D CNN model is trained using past wind data from the surrounding area of a wind farm, considering different time lengths. By analyzing the relationship between regional wind characteristics and the model's performance, we study the influence of spatiotemporal features of input wind data on predictive performances of CNN models.

Autocorrelation and Pearson correlation analyses are conducted on regional and seasonal wind data. Firstly, the autocorrelation analysis reveals that CNN predictive performances tend to decrease in regions with higher maximum standard deviations of the ACC across different regions. Notably, the regions with the highest maximum standard deviations are found to be in the following order: Korea, USA, and UK; while the predictive performances are observed to be the best in the reverse order: UK, USA, and Korea.

Secondly, Pearson correlation analysis is conducted to examine the spatial relationships within the data. Many low PCC values are observed in Korea, which align with the findings from the ACC analysis. The predictive performance of the CNN model increases when wind data from surrounding areas with high PCC values are included, while data from areas with low PCC values negatively impact the prediction performance. The variations in PCC value distribution among regions are attributed to meteorological and geographical factors, with Korea experiencing the most complex wind flow among the three regions due to these factors. Additionally, when training different CNN models (2D+3D-CNN model, Fully 3D-CNN model, and 1D-CNN model), it is observed that all of the CNN models exhibit a decrease in predictive performance when the spatiotemporal correlations of the regional wind are reduced.

In conclusion, incorporating favorably correlated spatial and temporal wind data from surrounding areas improves the predictive performances of CNN models. It is worth noting that the correlation analysis is a pure data analysis and does not involve training a CNN. Therefore, the proposed correlation analyses can be employed to estimate the learnability of a CNN prior to the training process. In addition, based on the findings in this study, it is recommended that wind turbines be installed in areas with high PCC values and small maximum standard deviations of ACCs to enable efficient power generation using CNN-based prediction/control methods.
\label{conclusions}

\section*{Acknowledgements} This work was supported by the National Research Foundation of Korea Grant funded by the Korean Government (NRF-2022R1F1A1066547); INHA UNIVERSITY Research Grant (2023). MR acknowledges the CoE RAISE project, which receives funding from the European Union’s Horizon 2020 –Research and Innovation Framework Programme H2020-INFRAEDI-2019-1 under grant agreement no. 951733. This work was performed as part of the Helmholtz School for Data Science in Life, Earth and Energy (HDS-LEE).

\end{linenumbers}
\bibliographystyle{elsarticle-num-names} 
\bibliography{references}

\end{document}